\documentclass[10pt,journal,twocolumn]{IEEEtran}

\usepackage{booktabs}
\usepackage{amsmath}
\usepackage{amsthm}
\usepackage{mathrsfs}
\usepackage{mathtools}
\usepackage{subfigure}
\usepackage{graphicx}
\usepackage{latexsym}
\usepackage{amssymb,amsbsy,verbatim,array}
\usepackage{epsfig}
\usepackage{epstopdf}
\usepackage{cite}
\usepackage{color}
\usepackage[ruled,lined,linesnumbered]{algorithm2e}
\usepackage{makecell}
\usepackage{extarrows}
\usepackage{multicol}
\usepackage{multirow}
\usepackage{bm}
\usepackage{mathrsfs}
\usepackage{booktabs}
\usepackage{siunitx}
\usepackage[switch]{lineno}
\usepackage{tabularx,booktabs}
\usepackage[flushleft]{threeparttable}
\usepackage[flushleft]{threeparttable}
\usepackage[short]{optidef}

\usepackage[colorlinks,
            linkcolor=red,       
            anchorcolor=blue,  
            citecolor=green,        
            ]{hyperref}

\begin{document}
%
\title{Computational Intelligence and Deep Learning for Next-Generation Edge-Enabled Industrial IoT}
%
%
%

\author{Shunpu Tang,
        Lunyuan Chen,
        Ke He,
        Junjuan Xia, Lisheng Fan, and Arumugam Nallanathan, \IEEEmembership{Fellow, IEEE}
 \thanks{S. Tang,  K. He, J. Xia and L. Fan are all with the School of Computer Science, Guangzhou University, Guangzhou 510006, China (e-mail: \{tangshunpu, heke2018\}@e.gzhu.edu.cn, \{xiajunjuan, lsfan\}@gzhu.edu.cn).}
  \thanks{L. Chen  is with the School of Electronics and Communication Engineering, Guangzhou University, Guangzhou 510006, China (e-mail: 2112019037@e.gzhu.edu.cn).}
  \thanks{A. Nallanathan is with the School of Electronic Engineering and Computer Science, Queen Mary University of London, London, U.K (e-mail: a.nallanathan@qmul.ac.uk).}
\thanks{S. Tang and L. Chen contributed equally to this work.}
\thanks{The corresponding author of this paper is L. Fan.}

}

%

\maketitle

\begin{abstract}
In this paper, we investigate how to deploy computational intelligence and deep learning (DL) in edge-enabled industrial IoT networks.  In this system, the IoT devices can collaboratively train a shared model without compromising data privacy. However, due to limited resources in the industrial IoT networks, including computational power, bandwidth, and channel state, it is challenging for many devices to accomplish local training and upload weights to the edge server in time. To address this issue, we propose a novel multi-exit-based federated edge learning (ME-FEEL) framework, where the deep model can be divided into several sub-models with different depths and output prediction from the exit in the corresponding sub-model. In this way, the devices with insufficient computational power can choose the earlier exits and avoid training the complete model, which can help reduce computational latency and enable devices to participate into aggregation as much as possible within a latency threshold. Moreover, we propose a greedy approach-based exit selection and bandwidth allocation algorithm to maximize the total number of exits in each communication round. Simulation experiments are conducted on the classical Fashion-MNIST dataset under a non-independent and identically distributed (non-IID) setting, and it shows that the proposed strategy outperforms the conventional FL. In particular, the proposed ME-FEEL can achieve an accuracy gain up to 32.7\%  in the industrial IoT networks with the severely limited resources.

\end{abstract}


%
\IEEEpeerreviewmaketitle
\section{Introduction}

With the advance of communication technologies and the emergence of new communication standards such as 5G and WiFi-6, more and more devices are connected to the Internet through wireless access points, which is called the era of Internet of Things (IoT)\cite{IoT}.
A large amount of data is generated and collected by many sensors and mobile devices, promoting the development of novel applications, including autonomous driving, augmented reality (AR) and smart cities.
However, a mass of data at the edge of the network brings great challenges to the central cloud servers.
Unfortunately, the conventional paradigm of cloud computing is unfriendly to those latency-sensitive applications.
To address these issues, mobile edge computing (MEC)\cite{MEC} was proposed to bring computational resources  close to the data source. In the concept of MEC, IoT devices can offload data and computational tasks to the edge server to achieve lower latency, lower energy consumption, and higher service quality.

In recent years, artificial intelligence (AI) technologies, e.g., machine learning (ML) and deep learning (DL), have made a breakthrough and even reached beyond human-level performances in image recognition, natural language processing, anomaly detection, and other domains\cite{DL}. To enhance the abilities of information processing and analysis, AI applications are deployed on edge devices to facilitate the emergence of edge intelligence\cite{edge_intelligence1,edge_intelligence2}. However, intelligent algorithms are usually computation- and energy-intensive. Meanwhile, limited computational power, battery capacity, and dynamic channel states restrict the application of edge intelligence.

Generally, collecting data and training models are the most two critical steps to deploy edge intelligence in IoT networks. Due to the decentralized nature of data in IoT networks, it is inappropriate to apply the conventional centralized learning in the cloud, which will increase the communication overhead. More importantly, uploading personal data may cause the leakage of privacy. In addition, many new and even fiercer laws were passed to prevent storing users' data on third-party servers.
To tackle the above issues, a collaborative training approach named federated learning (FL) was proposed to reduce the communication bandwidth, storage spaces, and energy consumption. In the FL,  the privacy can be protected by enabling multiple users to train a shared model by exchanging the weights instead of the sensitive raw data. It breaks away data islands and makes it possible to improve the performance of the data-driven model under the cooperation of multiple parties. However, FL still faces some challenges in edge-enabled IoT networks. One major challenge is system heterogeneity, where there exist different kinds of IoT devices with different computational power and channel state.

To coordinate and synchronize the FL training, a latency constraint should be set for all devices. The update will be discarded if the device can not accomplish uploading within the specified duration, which will cause performance degradation. Moreover, due to the limitations of computational power, channel state, battery capacity, and other factors, many devices often cannot train a deep model or upload weights successfully. This motivates us to design a novel framework of federated edge learning (FEEL), which enables devices to train a model with different scales under the limitations of latency and resources, and the server can still aggregate weights from different devices.

Inspired by the multi-exit mechanism proposed in \cite{anytime1}, this paper incorporates the multi-exit mechanism into FL and proposes the multi-exit-based federated edge learning (ME-FEEL). Specifically, we firstly add multiple exits in the deep model and enable it to output prediction from any exit. Therefore, the deep model can be divided into several sub-models, and devices can train parts instead of the whole deep model. Thanks to the multi-exit mechanism, the models trained by different devices still have the same sub-architecture that can be easily aggregated together. Moreover, a self-knowledge distilling (KD) approach is utilized to improve the performance of those early exits that lack of feature extraction and fitting abilities.   

Moreover, how to choose the best exit still remains an open problem to be solved. Although each device can choose an exit as deep as possible without violating the latency constraint, uploading time can not be neglected in bandwidth-limited  IoT networks, and it is usually a critical factor raising the failure aggregation. As described in \cite{FL_client, zzc_TVT,FEEL,Fl_client2}, it is useful to make the server aggregate the updates as much as possible with the given bandwidth allocation, promoting the training performance and stability. Thus, we aim to maximize the number of devices which upload their updates within the constrained latency. Notably, due to the use of multi-exit mechanism, an improved optimization objective is built to maximize the total number of exits which can upload successfully per FL round. We hence propose a heuristic exit selection and bandwidth allocation algorithm based on the greedy approach to solve the optimization problem.

In summary, this paper makes the following contributions.
\begin{itemize}
    \item We take into account the complicated scenario of edge intelligence, where there are heterogeneous devices with different computing power and channel state. The gap of computational power among devices can even reach tens of times, and the total communication bandwidth is limited. All devices will train a shared deep model collaboratively under a latency constraint.
    \item Inspired by the multi-exit mechanism, we propose a novel FL training framework named ME-FEEL, where devices with limited computational power can still participate into the FL by training a part of the deep model. Besides, a layer-wised model average strategy is presented to aggregate the models of different sizes. 
    \item Incorporating the limited bandwidth into IoT networks, we aim to maximize the number of devices which can successfully participate into the aggregation and propose an enhanced optimization objective that is to maximize the total number of exits per round. We also propose a joint exit selection and bandwidth allocation strategy based on the greedy approach to solve the problem.
    \item We conduct simulations and evaluate the proposed ME-FEEL by using a popular model and a practical dataset. Numerical results demonstrate that the proposed ME-FEEL can outperform the conventional strategies and it can be flexibly deployed in the edge-enabled industrial IoT networks.   
\end{itemize}

The rest of this paper is organized as follows. Section II briefly discusses related works on MEC and edge intelligence, and then Section III presents multi-exit FL and the associated workflow. Section IV further builds the exit selection and bandwidth allocation problem and provides the optimization algorithm. Simulations are performed in Section V to evaluate the proposed strategy. At last, the whole paper is concluded in Section VI.

\section{Related works}
The research of MEC has attracted much attention in recent years, and there are lots of relevant works about the offloading design and resource allocation \cite{chen_MEC,XUWEI_MEC,Zhou_MEC,Zichao_MEC,Xia_MEC}.
With the development of AI technologies, edge intelligence has become one of the most interesting topics. The definition of edge intelligence includes two aspects of \textit{AI for edge}  and \textit{AI on edge} \cite{Deng_EdgeAI}, which are detailed as follows.

Firstly, \textit{AI for edge} focuses on applying AI to deal with the complicated optimization problems, such as non-convex and NP-hard problems. For example, deep learning and reinforcement learning have achieved great success in the design of MEC networks, where a novel edge cache strategy based on deep Q-network (DQN) was proposed in \cite{Xiao_MEC}, and  
DRL and FL could be exerted  to optimize multiuser multi-CAP MEC networks \cite{Chao_MEC,GUO_FL}.

Moreover, \textit{AI on edge} focuses on the deployment of intelligent algorithms at the edge of network. To deal with the problem of limited resources in edge devices, efficient model architecture, model compressing approach and  hardware acceleration techniques should be proposed \cite{SqueezeNet,mobilenetv3,SongHan_Deep}.
From different perspectives, many researchers have designed some novel inference protocols. For example, the authors in \cite{Neurosurgeon} studied the characteristics of deep networks and presented an offloading strategy named Neurosurgeon, which could  automatically partition DNN computation between mobile devices and cloud servers. Besides, Zhao \emph{et. al} proposed a framework of distributed adaptive inference between edge devices and servers to minimize the memory footprint \cite{Zhao_Deepthings}. In further, as a distributed DNN computing system, CoEdge was proposed to orchestrate cooperative inference over heterogeneous IoT devices and helped reduce the energy consumption  up to $25\%$  \cite{CoEdge}.

Concentrating on deployment of FL in edge-computing based IoT networks, researchers have made many efforts in the past few years. In this field, the authors in \cite{Fl_client2} solved the problem of client selection in resource-constraint mobile edge computing networks. To address the problem of communication overhead, the authors in \cite{chen_FL} took  into account the difference between the uplink and downlink communications, and proposed a physical-layer quantization strategy. Moreover, the bandwidth allocation could be optimized to reduce the system cost, by using an intelligent swarm algorithm \cite{zzc_TVT}. For the heterogeneous system, some intelligent scheduling approaches could be developed to enable the devices which were not the system bottleneck to reduce unnecessary energy consumption \cite{Zhan_FL}. In further, the authors in \cite{FEEL} studied the
IoT devices powered by different batteries, and proposed a DDPG-based approach to prolong the battery life.

\begin{figure}[t!]
    \centering
    \includegraphics[scale=0.4]{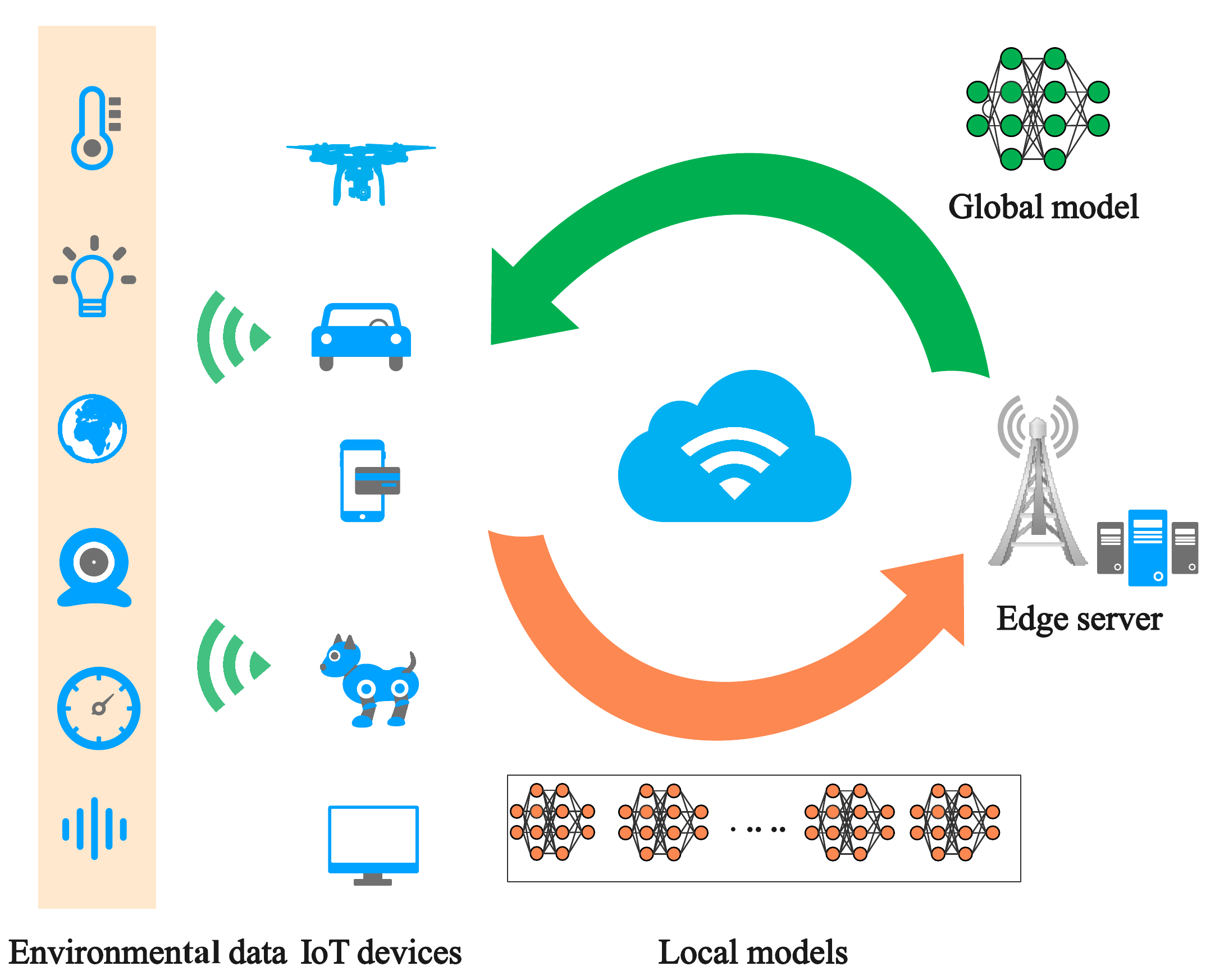}
    \caption{Federated edge learning system.}
     \label{Fig::Iot}
\end{figure}

\section{Multi-exit based federated learning}
In this section, we investigate the FEEL system in IoT networks, which consists of one edge server and multiple IoT devices. We then present the key design of the proposed ME-FEEL by giving the formulation and working mechanism of multiple exits. We further explore the aggregation strategy for the models of different sizes. 

\subsection{Federated edge learning system}
Fig. \ref{Fig::Iot} depicts the system model of federated edge learning, where there is one edge server and $K$ IoT devices.  For IoT device $k$ with $1\leq k \leq K$, it firstly collects the local dataset $\mathcal{D}_k=\{(\bm{x}_n,y_n)|1\leq n \leq N_k\}$, where $N_k$ is the number of the local training samples in device $k$, while $\bm{x}_n$ and  $y_n$ denote the training sample and label, respectively.  These $K$ devices will train a shared deep model with the help of the edge server, which aims to minimize the global objective function $f(w)$, which can be given by
\begin{align}
    \min_{w} f(w)=\sum_{k=1}^{K}\frac{N_k}{N}F_k(w),
\end{align}
where $w$ is the model weights,  $F_k$($ \cdot $) represents the local loss function, and $N=\sum_{k=1}^K N_k$ is the total number of samples. For device $k$, it performs the training on its local dataset in parallel, and updates the weights by the gradient descent algorithm,
\begin{align}
    w_k \xleftarrow{}w_k-\eta \nabla F_k(w_k),
    \label{Eq::sgd}
\end{align}
where $\eta$ is a learning rate. In practice, to reduce the commutation overhead, the edge server randomly chooses one device subset $\mathcal{K}$ in per round, where $|\mathcal{K}|\ll K$. Then, the  chosen device $k\in \mathcal{K}$ performs the gradient descent algorithm $e$ times. After that, to optimize the global objective function, the local update will be uploaded to the edge server, which can aggregate the weights from different devices and obtain a global update by
\begin{align}
    w^{r+1}=\frac{1}{|\mathcal{K}|}\sum_{k\in \mathcal{K}}w_k^{r+1},
\end{align}
where $r$ and $r+1$ are the indices of the current and next training rounds, respectively. The overall procedure of FEEL is summarized in Algorithm \ref{Algo::fedavg}.
\begin{algorithm}[t!]
    \textbf{Input} $K$, $R$, $\eta$ $w^0$\;
    \For{\text{Round} $r=0,\cdots, R-1 $}{
        Server randomly chooses a subset $\mathcal{K}$ from $K$ devices\;
        Each device $k\in \mathcal{K}$ downloads the global weights $w^r$ from the server\;
        Each device $k\in \mathcal{K}$ obtains the weights $w_k^{r+1}$ by performing SGD with a learning rate $\eta$ on the local dataset\;
        Each device uploads the weights $w^{r+1}_k$ to the server\;
        The edge server aggregates the weights as $w^{r+1}=\frac{1}{|\mathcal{K}|}\sum_{k\in \mathcal{K}}w_k^{r+1}$
    }
    \caption{FEEL procedure}\label{Algo::fedavg}
\end{algorithm}





\subsection{The proposed multi-exit approach}
To tackle the system heterogeneity in IoT networks, we propose a novel FEEL training framework named ME-FEEL,
which enables the chosen devices to participate into training to the best of their abilities by setting multiple exits for the deep models.

\begin{figure}[t!]
    \centering
    \includegraphics[scale=0.6]{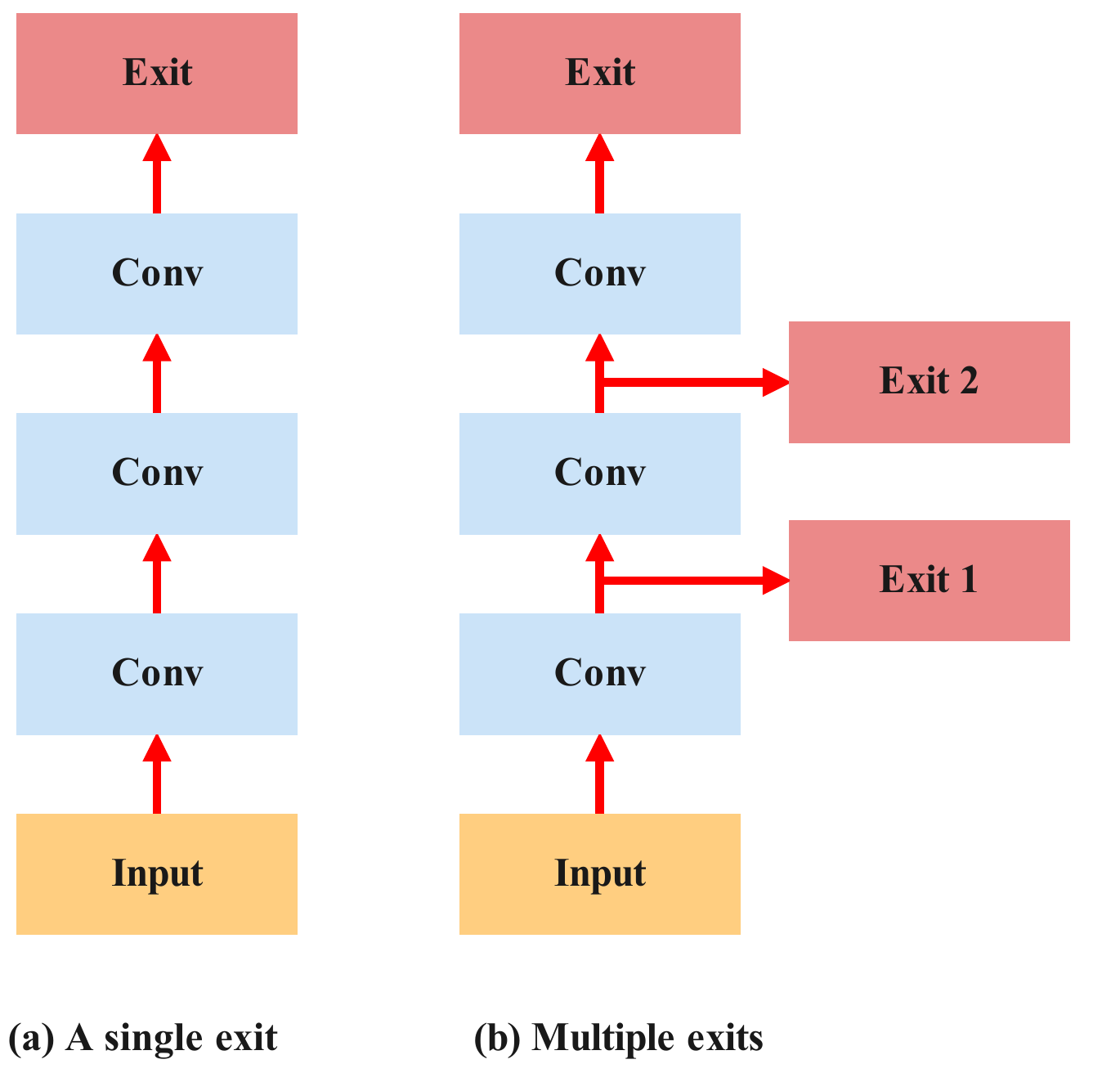}
    \caption{An example of multi-exit mechanism.}
     \label{Fig::M_exit}
\end{figure}

 As shown in Fig. \ref{Fig::M_exit} (a), in the original deep model such as a fully connected layer based classifier, there is only one exit which is usually  the last layer. The features of input will be extracted by continuous convolution kernels and then be fed into the classifier. The multi-exit architecture is illustrated in Fig. \ref{Fig::M_exit} (b), where there are multiple exits at different depths. In contrast, it is not necessary to wait until the computation in the last layer is completed to obtain the prediction result. All exits share a part of weights, and there are not many additional parameters added in the model. This makes the model separable and can be divided into multiple sub-models. Obviously, deeper exits cause more resource consumption and a larger latency. Formally, the output of the model with $M$ exits can be expressed as
\begin{align}
    \mathcal{P}=\{\bm{p}_1,\cdots \bm{p}_M\},
\end{align}
where $\bm{p}_m \in \mathcal{P}$ denotes the output logits of the $m$-th exit, and we use $p_{m}(x)$ to denote the mapping function from the input $x\in \mathcal{X}$ to $\bm{{p}_m}$. In practice, device $k$ can adaptively choose $M_k \leq M$ and train the continuous part $p_1(x), \cdots p_{M_k}(x)$, depending on its computational power and channel state. Thus, it can also quickly adapt to the dynamic environment in the IoT networks.

\begin{figure}[t!]
    \centering
    \includegraphics[scale=0.34]{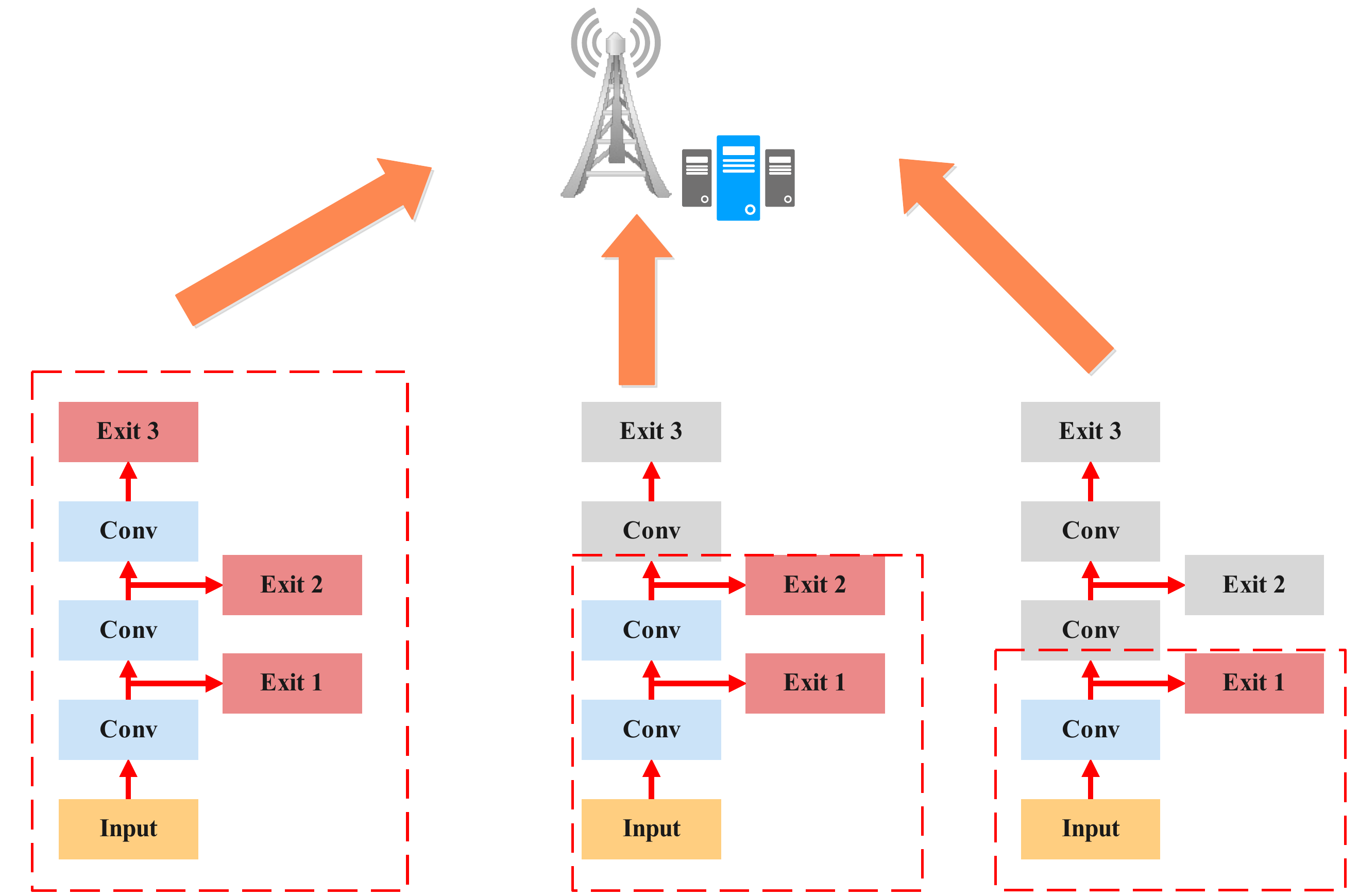}
    \caption{Training framework of ME-FEEL.}
     \label{Fig::exit_FL}
\end{figure}

We further design the training framework of ME-FEEL shown in Fig. \ref{Fig::exit_FL}, and it consists of the following the pivotal steps,
\subsubsection{Local training} For an arbitrary device $k$, it can choose $M_k$ exits according to its current state. Given the local dataset $\mathcal{D}_k$, the local model can output $M_k$ results for each sample. Thus, to train all exits and shared weights, the local loss function on dataset $\mathcal{D}_k$ can be regarded as the sum loss of all exits, which can be written as
\begin{align}
     F_{k}(w;\mathcal{D}_k)=\frac{1}{N_k} \sum_{n=1}^{N_k} \mathcal{L}_{pre}(y_n,p_m(\bm{x}_n)),
     \label{Eq:local}
\end{align}
where
\begin{align}
    \mathcal{L}_{pre}(y,p_m(\bm{x}))=\frac{1}{M_{k}} \sum_{m=1}^{M_k} \ell_{pre}(y,p_m(\bm{x})),
\end{align}
in which $\ell_{pre}(\cdot)$ represents the mean square error of the regression task and cross-entropy for classification tasks.

In general, the later exit outperforms the earlier one, because a deeper layer can often extract more abstract features and exhibit a stronger representational ability. In other words, the later exits often have more knowledge than the earlier ones. Hence, the earlier exits are hard to train well from scratch and do not have enough ability to process difficult samples properly. From the viewpoint of knowledge distilling (KD) \cite{knowledge_distilling_Hinton}, the exits with more knowledge can be viewed as teachers which can transfer their knowledge to those students with less knowledge in order to help improve the accuracy\cite{anytime1}. Therefore, each device can employ the KD-based strategy to train the multiple exits. The objective of KD is to minimize the  difference between the output distributions of the teacher and students. Formally, we use Kullback-Leibler (KL) divergence  to describe the difference, which can be written as
\begin{align}
    \ell_{kd}(\bm{s},\bm{t})= -\tau^2 \sum_i z_i({\tau}) \log v_i({\tau}),
    \label{Eq:kd_loss}
\end{align}
where $\bm{s}$ and $\bm{t}$ are the output logits of the students and teacher, respectively. Notation $\tau$ is the temperature parameter of KD, and $v_i({\tau})$
can be represented as
\begin{align}
 v_i({\tau})=\frac{e^{s_i/\tau}}{\sum_j e^{s_j/\tau}}.
 \label{Eq:soft1}
\end{align}
Analogously, notation $z_i{(\tau)}$ is defined as
\begin{align}
 z_i({\tau})=\frac{e^{t_i/\tau}}{\sum_j e^{t_j/\tau}}.
  \label{Eq:soft2}
\end{align}
We can use the temperature parameter $\tau$ to soften the output of the teacher\cite{knowledge_distilling_Hinton}. In particular, when $\tau=1$,  (\ref{Eq:soft1})-(\ref{Eq:soft2}) degrade into the SoftMax function, and it is easy to ignore the information from the negative samples. On the contrary, a higher temperature with $\tau > 1$ can avoid the overconfidence of neural network and it can allow the output to contain more information on the similarity of different classes. Moreover, it is important to multiply $\tau^2$ in (\ref{Eq:kd_loss}) to ensure that the change of temperature in the experiment will not affect the magnitude of gradient.

In the aforementioned model with $M$ exits,  we further regard the teacher $\bm{t}(\bm{x})$ as the ensemble output of all exits, which can be written as 
\begin{align}
    \bm{t}(\bm{x})=\frac{1}{M_{k}}\sum_{m=1}^{M_{k}}p_m(\bm{x}).
     \label{Eq::teacher}
\end{align}
This is because that the ensemble results have stronger generalization performance than a single result, which can reduce the possibility of overfitting. 
Hence, the total KD loss can be computed as
\begin{align}
    \mathcal{L}_{kd}(\bm{x})=\frac{1}{M_{k}} \sum_{m=1}^{M_k} \ell_{kd}(p_m(x),\bm{t}(\bm{x})).
\end{align}
By jointly using the prediction loss and KD loss, (\ref{Eq:local}) can be rewritten
\begin{align}
     F_{k}(w;\mathcal{D}_k)=\frac{1}{N_k}\sum_{n=1}^{N_k} \left[ \mathcal{L}_{pre}(y,p_m(\bm{x}_n)+ \mathcal{L}_{kd}(\bm{x}_n)\right],
     \label{Eq::joint_loss}
\end{align}
which will be minimized by the local gradient descent.
\subsubsection{Model aggregation}
After each device  completes the local update in parallel, the updated weights will be uploaded to the edge server. In the conventional FEEL, the updated weights of different devices are of the same dimension. Therefore, model average should be modified to adapt the architecture of ME-FEEL.

To tackle this problem, a simple but efficient strategy is to aggregate the shared parts of different models. Although the upload weights are of different sizes, they all have the same sub-architecture. Specifically, each layer in the model with fewer exits can be found in the model with more exits. Thus, we propose a layer-wised model average strategy to obtain the global model, namely multi-exit FedAvg (ME-FedAvg). The edge server will take the initial global model as a reference, and then perform the search layer by layer in the uploaded updates. If one layer is found to exist in the model uploaded by some devices, then the edge server will average the weights of the layer from different devices.

Mathematically, we suppose that there are $L$ layers with parameters in the initial global model, and $L_k$ is used to denote the number of layers in the $k$-th device. For each layer $l\in [1,L]$ in the global model, its weight $w(l)$  can be updated by
\begin{align}
w{(l)}=\frac{\sum_{c \in \mathbb{C}_l}{|D_c| w_c{(l)}}}{\sum_{c \in \mathbb{C}_l}{|D_c|}}, \forall l\in[1,L],
\end{align}
where $\mathbb{C}_l$ is a subset of $\mathcal{K}$. The edge server searches for devices with the $l$-th layer, and then produces the device subset $\mathbb{C}_l$. It means that all the devices in $\mathbb{C}_l$ have layer $l$. In other words, the number of layers in the device $ c \in \mathbb{C}_l$ is greater than the value of $l$, which can be written as
\begin{align}
L_c \geq l, \forall c \in \mathbb{C}_l.
\end{align}

To summarize, we provide the procedure of the proposed ME-FEEL in Algorithm \ref{Algo::ME-FEEL}. Different from the convention FL, we set multiple exits in the deep model, and enable devices to train the model with different sizes. We will further propose a joint exit selection and bandwidth allocation strategy as shown in line 5 of Algorithm \ref{Algo::ME-FEEL}, which is detailed in the following section.

\begin{algorithm}[t!]
    \textbf{Input} $K$, $T$, $\eta$ $w^0$, $L$\;
    \For{\text{Round} $t=0,\cdots, T-1 $}{
        The edge server randomly chooses a subset $\mathcal{K}$ from $K$ devices\;
        Each device $\in \mathcal{K}$ downloads the global weights $w^t$ from the server\;
        Device $k$ chooses $M_k$  exits and the MEC controller allocates bandwidth to the device according to the current environment\;
        Device $k$ obtains the weights $w_k^{t+1}$ by performing SGD on  (\ref{Eq::joint_loss}) with the learning rate $\eta$\;
        Device $k$ uploads the weights $w^{t+1}_k$ to the server\;
        \For{$l=1, \cdots L$}{
        The edge server produces the subset of devices $\mathbb{C}_l$\;
        The edge server aggregates the weights of the $l$-th layer as $w^{t+1}(l)=\frac{\sum_{c \in \mathbb{C}_l}{|D_c| w_c{(l)}}}{\sum_{c \in \mathbb{C}_l}{|D_c|}}$\;
        }
    }
    \caption{Multi-exit FEEL}\label{Algo::ME-FEEL}
\end{algorithm}

\section{Exit selection and bandwidth allocation}
In this section, we present how to choose proper exit points and propose a joint exit selection and bandwidth allocation strategy, which can adapt the dynamic environment in the IoT networks. Specifically, we firstly describe the latency constrained model of FEEL, and then provide the problem formulation. We further propose a heuristic algorithm based on the greedy approach to solve the optimization problem.

\subsection{Latency constrained model}
As mentioned before, all devices train the shared model under a latency constraint denoted by $\gamma_{th}$. The total communication bandwidth is $B_A$, and it is managed by the MEC controller. Therefore, for each device in $\mathcal{K}$, it uploads its weights through wireless channel within $\gamma_{th}$, i.e.,
\begin{align}
    T_{\text{local},k}+T_{\text{up},k}\leq \gamma_{th},
    \label{Eq:time_constrain}
\end{align}
where $T_{\text{local},k}$ and $T_{\text{up},k}$ are the latency of local training and data transmission, respectively. The local latency depends on the number of exits and the number of samples in the local dataset $D_k$, which can be characterised by
\begin{align}
    T_{\text{local},k}=\frac{\alpha_{k} |D_k|\cdot g_1(M_k)}{C_k},
\end{align}
where $\alpha_{k}>0$ is a coefficient of computational power, $C_k$ is the batch size of devices $k$, and $g_1(\cdot)$ represents the mapping from the number of exits to the training time of only one batch. 

According to the Shannon theorem, the transmission data rate between device $k$ and BS is
\begin{align}
R_{k}=B_k \log_2 \Bigg ( 1+\frac{P_k|h_k|^2}{\sigma_k^2}\Bigg ),
\end{align}
where $B_k$ is the allocated channel bandwidth, $P_k$ denotes the transmit power, $h_k$ is the channel parameter, and $\sigma_k^2$ is the variance of the additive white Gaussian noise (AWGN). Therefore, the transmission latency is given by
\begin{align}
    T_{\text{up},k}=\frac{g_2(M_k)}{R_k},
    \label{Eq::t_up}
\end{align}
where $g_2(\cdot)$ is the mapping from the number of exits to the number of bits.

\subsection{Problem formulation}
In the FEEL system, it is useful to make the server aggregate updates as much as possible through reasonable resource management and scheduling, which can help improve training performance and stability. Similarly, in the proposed ME-FEEL, the number of exits also affects the learning performance. Thus, we aim to maximize the number of exits that can upload their updates without violating the latency threshold $\gamma_t$.

Mathematical, we use $\mathcal{K}_s$ to denote the device set in which the devices can upload their updates successfully, where $\mathcal{K}_s \subseteq \mathcal{K}$. Further, the problem formulation is give by 

\begin{maxi*}
        {\{{M_k},{B_k}\}}{\sum_{k \in \mathcal{K}_s}M_k \tag{P1}\label{P1}}
        {}{}
        \addConstraint{T_{\text{local},k}+T_{\text{up},k}\leq \gamma_{th}\tag{c1}}
        \addConstraint{\sum_{k\in \mathcal{K}_s}B_k \leq B_A.\tag{c2}}{}{}
   \end{maxi*}
 where $B_A$ is the total communication  bandwidth.
\subsection{Optimization strategy}
Note that the problem in (P1) is a variant of classical knapsack problem (KP) and it belongs to the NP-complete problem\cite{KP}. It has been pointed out in the literature that the (P1) can be solved at the cost of pseudo-polynomial time. To facilitate solve this problem, we propose a heuristic exit selection and bandwidth allocation algorithm based on the greedy approach, detailed as follow. 

From (\ref{Eq:time_constrain})-(\ref{Eq::t_up}), the bandwidth $B_{k}(m)$ for uploading $m$ exits with the constraint c1 should meet the following requirement
\begin{align}
    \frac{g_2(m)}{B_{k}(m) \log_2 \Bigg ( 1+\frac{P_k|h_k|^2}{\sigma_k^2}\Bigg )}+\frac{\alpha_{k} |D_k|\cdot g_1(m)}{C_k} \leq \gamma_{th},
\end{align}
where $m\in [1,M]$ is the index of exits. We then have
\begin{align}
   B_{k}(m) \geq \frac{g_2(m)}{\Bigg (\gamma_{th}-\frac{\alpha_{k} \left|\mathcal{D}_{k}\right| g_1(m)}{C_{k}}\Bigg ) \log _{2}\Bigg (1+\frac{P_{k}\left|h_{k}\right|^{2}}{\sigma_{k}^{2}} \Bigg )}. \label{Eq::required_b}
\end{align}
 Therefore, we can set $ B_{k}(m)$ to the minimum required bandwidth for the $m$-th exit, 
 \begin{align}
   B_{k}(m) \triangleq \frac{g_2(m)}{\Bigg(\gamma_{th}-\frac{\alpha_{k} \left|\mathcal{D}_{k}\right| g_1(m)}{C_{k}}\Bigg) \log _{2}\Bigg(1+\frac{P_{k}\left|h_{k}\right|^{2}}{\sigma_{k}^{2}}\Bigg)}. \label{Eq::required_b_2}
\end{align}
\begin{algorithm}[t!]
    \textbf{Input} $K$, $B_A$, $M$\;
     \For{$k\in \mathcal{K}$}{
       $m=M$\;

        \tcp{ Computing the required bandwidth according to (\ref{Eq::required_b})}
     \While{$B(m,k)\leq 0$ \mbox{or} $m>0$}
     {

        $m=m-1$\;
     }
     \If{$m \textgreater 0$}
     {Device $k$ chooses exit $M_k=m$\;
     Device $k$ sets bandwidth to $B_k=B_{k}(M_k)$\;
     Add device $k$ to $\mathcal{K}_s\;$
     }
     }
      \tcp{Adjust the exit selection}
     $B=\sum_{k\in \mathcal{K}_s}B_k$\;
     \While{$B\leq B_A$}{
     $k_{min}=\arg \min_{k\in \mathcal{K}_s} \frac{M_k}{B_k}$\;
     $M_{k_{min}}= M_{k_{min}}-1$\;
     $B_{k_{min}}=B_{k_{min}}(M_{k_{min}})$\;
     $B=\sum_{k\in \mathcal{K}_s}B_k$\;
     }

     \textbf{Output} \{$B_k$, $M_k$\}


    \caption{Exit selection and bandwidth allocation}\label{Exit selection}
\end{algorithm}
 The key idea of making decision is to adjust the exit according to the radio of $M_K$ and the minimum required bandwidth $B_{k}(m)$ until the bandwidth constraint is met. As described in Algorithm \ref{Exit selection}, each device firstly checks whether it can accomplish the training within $\gamma_{th}$ from the latest exit to the first one ($M$ to $1$) by computing $B_{k}(m)$ (lines 2-6). Specifically, $B_{k}(m) \leq 0$ represents that the device fails to finish the local training, due to the non-positive item of  $\gamma_{th}-\frac{\alpha \left|\mathcal{D}_{k}\right| g_1(m)}{C_{k}}$. In contrast, $B_{k}(m) > 0$ represents that the device can accomplish the local training within the latency threshold $\gamma_{th}$. Secondly, after each device preliminarily chooses a proper exit, it will report this information to the MEC controller and apply for the required bandwidth (lines 7-10).  After that, the MEC controller will play a very important role, which takes charge of global scheduling. In line 13, the MEC controller will integrate the information from different devices and compute the current bandwidth requirement. If the bandwidth constraint is not satisfied, the edge server will fine-tune the exit of device in a greedy approach. As shown in line 14-18, the edge server finds the device $k_{min}$ whose radio of $M_K$ and $B_{k}(m)$ is the minimum, and then sets the exit forward as $M_{k_{min}}=M_{k_{min}}-1$. Lastly, this process of exit adjustment will repeat until without violating the bandwidth constraint.


\section{Simulation and Discussion}
In this section, we conduct simulations to evaluate the performance of the proposed ME-FEEL. Specifically, we  firstly introduce the simulation setting and the implementation details, and then we will present the simulation results and discussion.
\subsection{Simulation Settings}
We conduct the simulation environment of edge intelligence in the IoT network, consisting of one edge server on the BS and 100 edge devices. These devices have their own datasets and can participate into the FL. The setup details are presented below.
\begin{figure}[t!]
    \centering
    \includegraphics[scale=1.0]{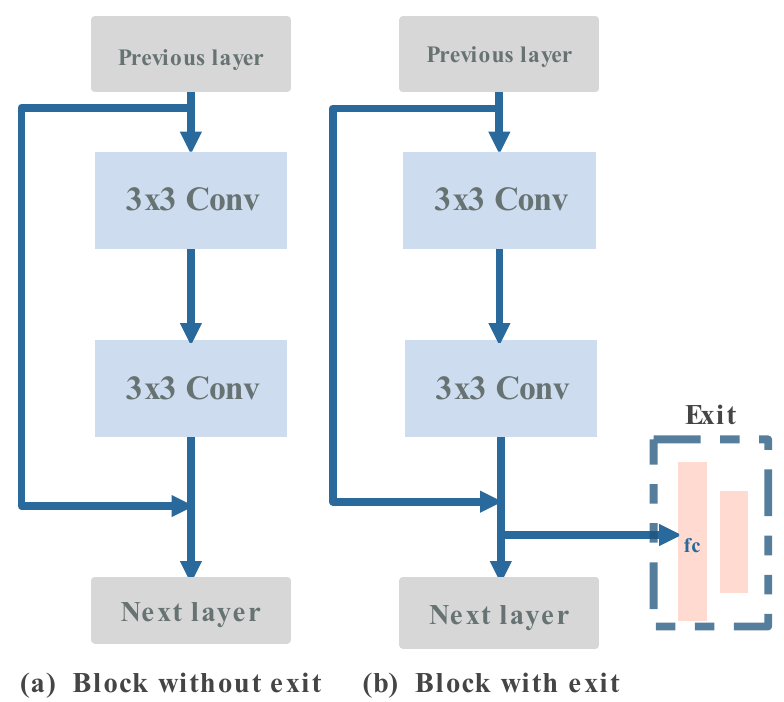}
    \caption{Block architecture in the ME-ResNet.}
     \label{Fig::block}
\end{figure}
\subsubsection{Task and Dataset}
We consider the common application on image classification, and the devices train a shared classification model collaboratively. Specifically, a typical image dataset Fashion MNIST\footnote{This dataset can be found through the link \url{https://github.com/zalandoresearch/fashion-mnist}.}
with 10 output classes is used in the simulation. The dataset has 5000 training samples and 1000 test samples, and they will be divided into 100 parts. The training samples are partitioned with the same non-IID setting \cite{FL_McMahan}. We firstly sort the data by the index of classes and then divide them into 200 groups of size 300, ensuring that only two groups are assigned to one device.

\begin{table}[h]
    \centering
     \caption{The modified ResNet-18 with 7 exits.}
    \label{tab:ME-resnet}

   \begin{tabular}{cccccc} \toprule
    {Input} & {Operator} & {Channels} & {Stride} & {Output} & Exit\\ \midrule
    ${32}^2 \times 1$  & Conv$3\times 3$ & 64 & 1 &${32}^2 \times 64$ & - \\\midrule
    ${32}^2 \times 64$  & Block & 64 & 1 &${32}^2 \times 64$ & - \\
    ${32}^2 \times 64$  & Block & 64 & 1 &${32}^2 \times 64$ & \checkmark \\\midrule
    ${32}^2 \times 64$  & Block & 128 & 2 &${16}^2 \times 128$ & \checkmark \\
    ${16}^2 \times 128$  & Block & 128 & 1 &${16}^2 \times 128$ & \checkmark \\ \midrule
    ${16}^2 \times 128$  & Block & 256 & 2 &${8}^2 \times 256$ & \checkmark \\
    ${8}^2 \times 256$  & Block & 256 & 1 &${8}^2 \times 256$ & \checkmark \\ \midrule
    ${8}^2 \times 256$  & Block & 512 & 2 &${4}^2 \times 512$ & \checkmark \\
    ${4}^2 \times 512$  & Block & 512 & 1 &${4}^2 \times 512$ & - \\ \midrule
    ${1}^2 \times 512$  & FC & - & - &$10$ & \checkmark \\
    \bottomrule
\end{tabular}

\end{table}
\subsubsection{Multi-exit FL}
We use the classical ResNet-18\cite{resnet} due to its outstanding feature extraction ability. We modify the first layer to make it adapt the input size. 
Besides, the original building block consists of two continuous $3\times 3$ convolutions and one shortcut. As shown in Fig. \ref{Fig::block}, we add an exit on the bottom of the block, where the exit includes a full-connected network to output the prediction. Similarly, $7$ exits in total are set in ResNet-18, and the complete model architecture named ME-ResNet is shown in Table \ref{tab:ME-resnet}. Notably, necessary activate functions and reshape operations are omitted for simplification. As well, all the batch normalization (BN) layers are replaced by group normalization for improving the performance under non-IID setting. We use Adam optimizer to update weights with a learning rate of 0.001. Batch size $C_k$ is 10, and we set the temperature of KD to 3. The total number of communication rounds is 750, and only the chosen 10 devices will train the local model for 5 times in each round. Another important setting is the mappings $g_1(\cdot)$ and $g_2(\cdot)$, where we run practical measurement on the AMD workstation, and the results are  presented later.

\subsubsection{Network environment}
For the wireless communication system in the IoT network, we set up all devices to connect to the base station over LTE cellular networks. The available bandwidth of BS is 40MHz and the transmit power of each device is set to 1W. To simulate the variable channel state, $|h_k|^2$ is subject to the exponential distribution with the average channel gain of 1. It remains the same for each round and while varies between different rounds. The variance of AWGN is set to $1e-3$. 

\subsection{Competitive methods}
In order to verify the effectiveness of the proposed strategy, we compare it with various methods of FEEL listed below. 
\begin{itemize}
    \item \textbf{FEEL-ideal}: The conventional FEEL algorithms in the ideal IoT network, where no constraint is considered at all.
    \item \textbf{FEEL}: Based on the FEEL-ideal, we incorporate the practical IoT network under latency and bandwidth constraint. For the limited bandwidth, we apply the bandwidth allocation method introduced in \cite{zzc_TVT}, where the MEC controller always allocates the bandwidth firstly to the clients requiring less.
    \item\textbf{FEEL-UB}: Different from the FEEL, the edge server allocates the bandwidth evenly among the devices.
\end{itemize}
\subsection{Simulation results and discussion}
\begin{figure}[t!]
    \centering
    \includegraphics[width=3.5in]{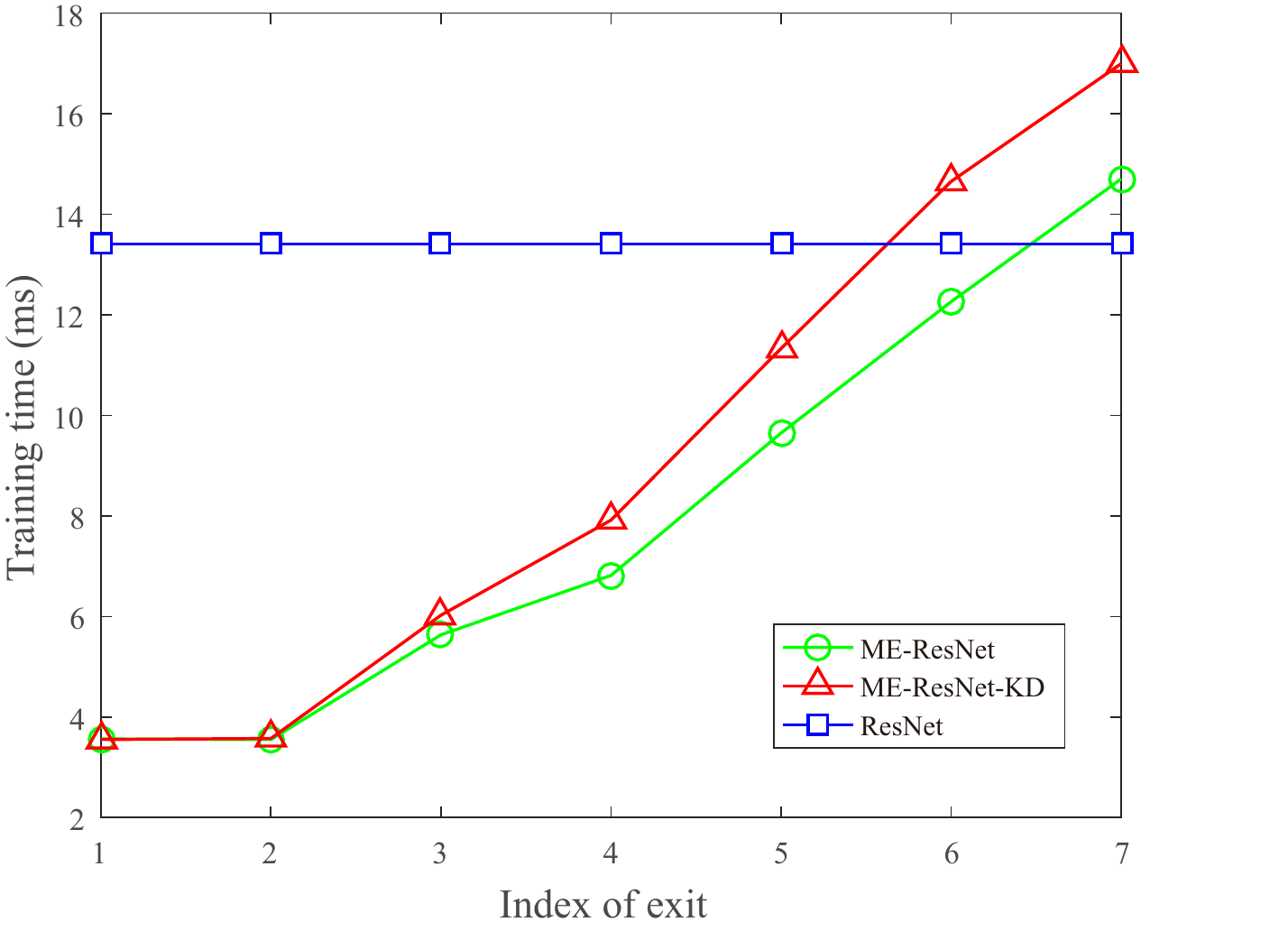}
    \caption{Local training time versus the index of exit.}
     \label{Fig::training_time}
\end{figure}
\begin{figure}[t!]
    \centering
    \includegraphics[width=3.5in]{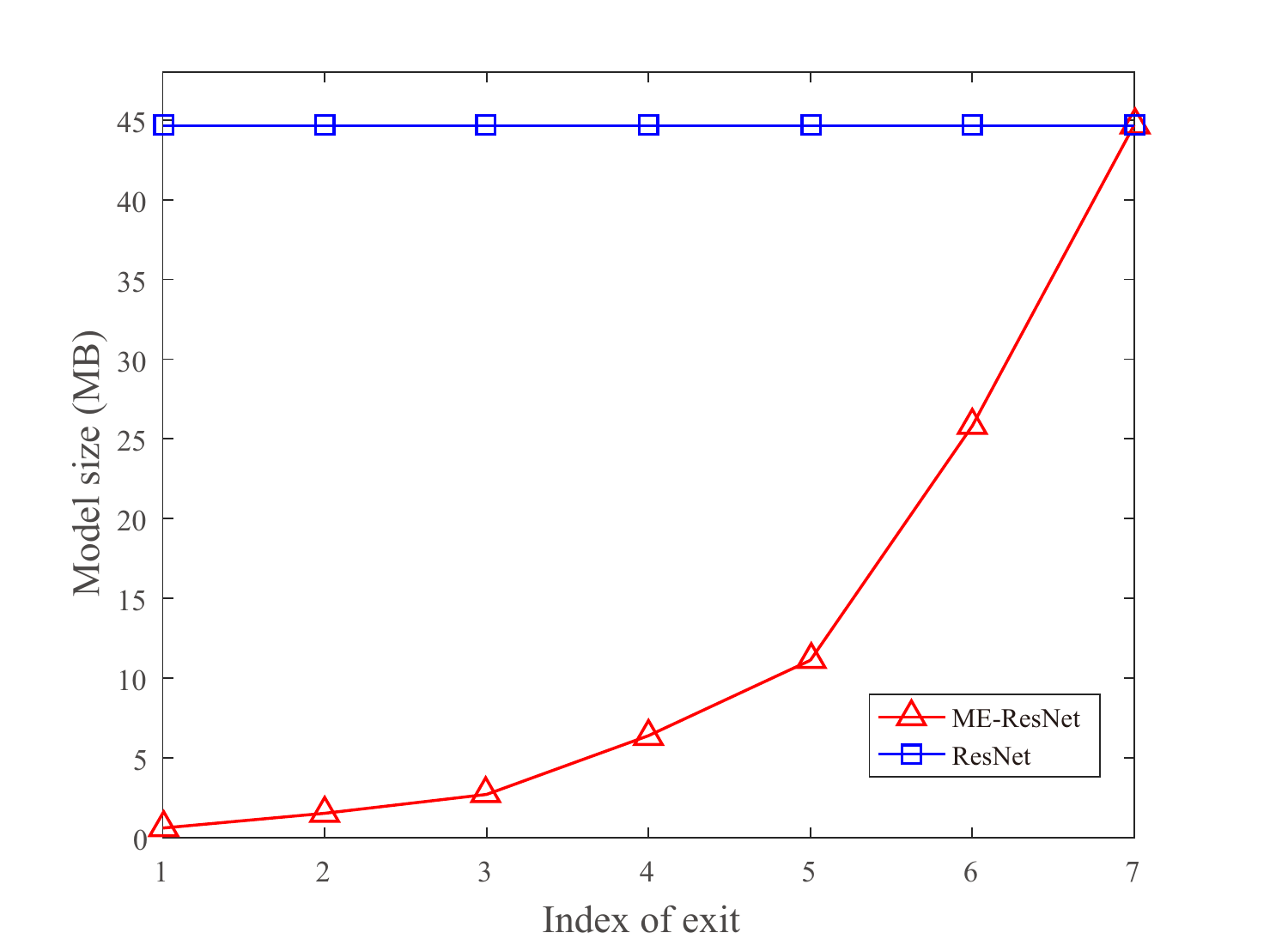}
    \caption{Model size in MBytes versus the index of exit.}
     \label{Fig::model_size}
\end{figure}

\begin{figure*}[t]
    \centering
    \includegraphics[width=7.2in]{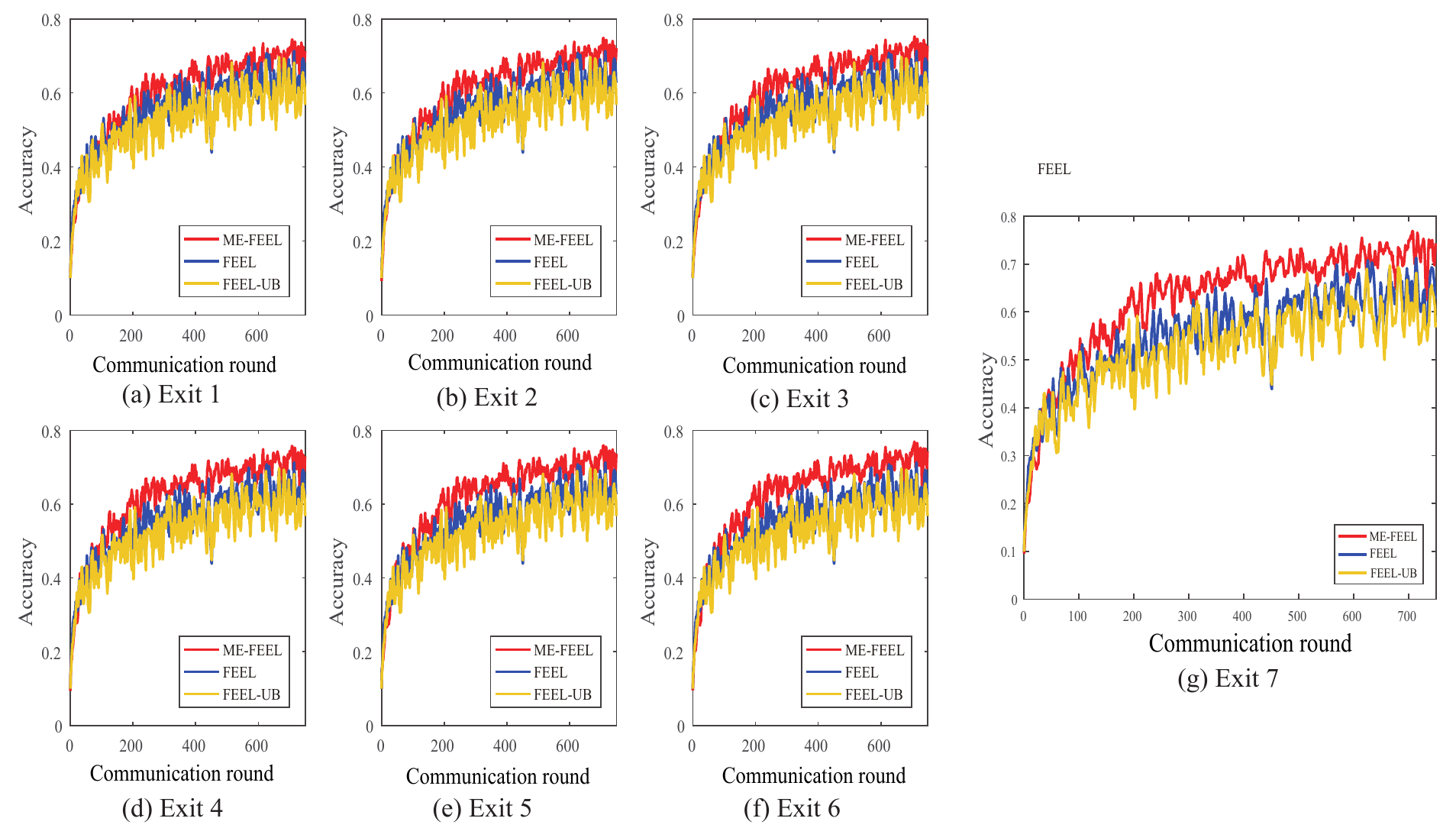}
    \caption{Test accuracy versus the communication round for the Fashion-MNIST dataset  with $\gamma_{th}=15$s and $B_A=40$MHz.}
     \label{Fig::acc_vs_rounds}
\end{figure*}
\begin{figure}[t]
    \centering
    \includegraphics[width=3.5in]{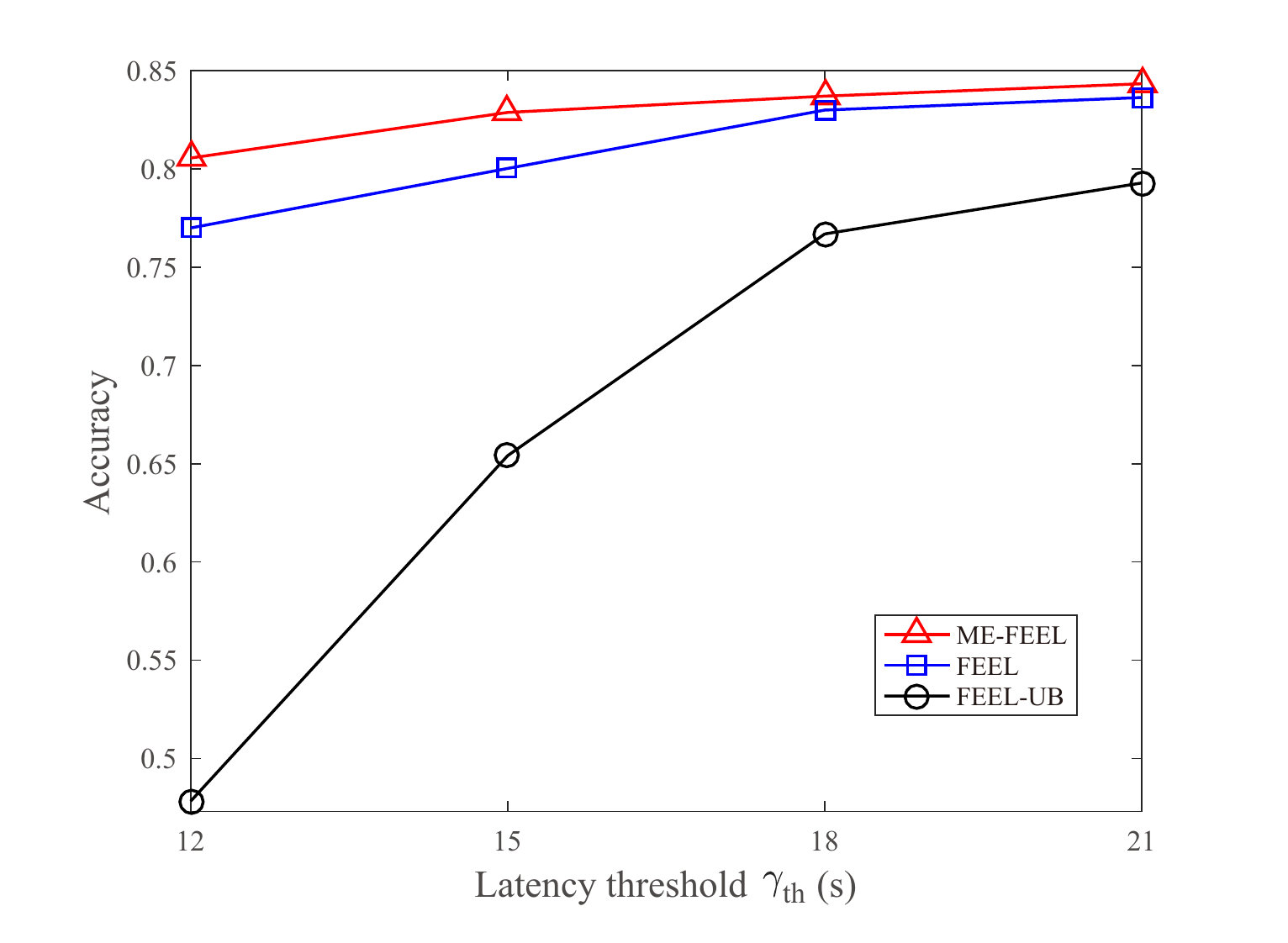}
    \caption{Test accuracy versus the latency threshold $\gamma_{th}$ for the Fashion-MNIST dataset  with $B_A=40$MHz.}
     \label{Fig::acc_vs_time}
\end{figure}
\begin{figure}[t]
    \centering
    \includegraphics[width=3.5in]{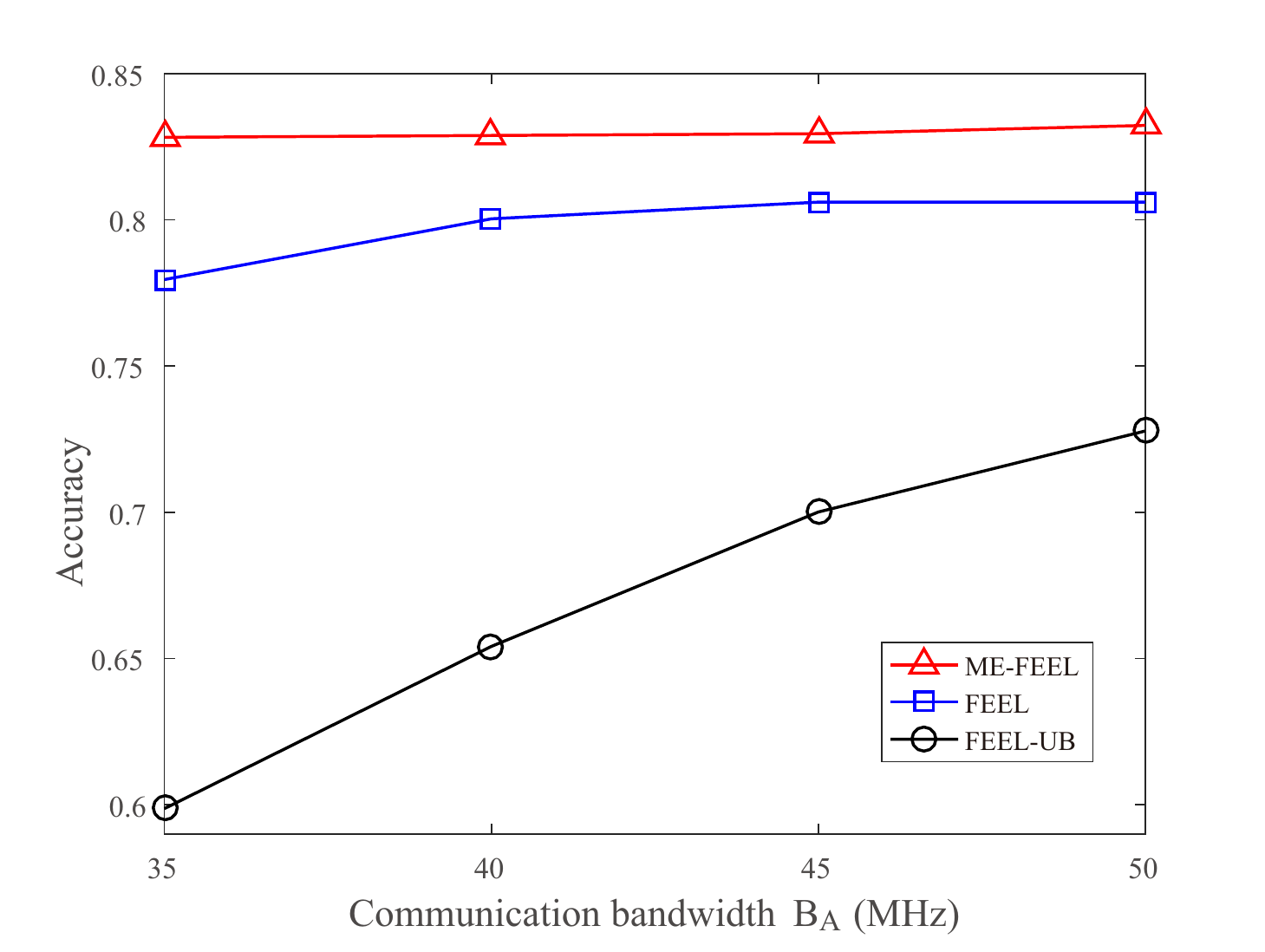}
    \caption{Test accuracy versus the total communication bandwidth $B_{A}$ for the Fashion-MNIST dataset  with latency threshold $\gamma_{th}=15$s.}
     \label{Fig::acc_vs_bandwidth}
\end{figure}

Fig. \ref{Fig::training_time} depicts the local training time versus the index of exit of the original ResNet, ME-ResNet, and ME-ResNet with KD technology (ME-ResNet-KD), where the number of exits is 7 in ME-ResNet, and the number of training samples is 10. From this figure, we can find that the original ResNet with only one exit requires about 14ms to train a batch of samples. When the exit is the last one, the training time of ME-ResNet-KD and ME-ResNet is 16.9ms and 14.70ms, respectively. However, when we choose some earlier exits, the training time of ME-ResNet and ME-ResNet-KD can be reduced significantly. In particular, compared with the ResNet, ME-ResNet and ME-ResNet-KD can reduce the training time to half when  the $4$-th exit is chosen. Moreover, when the ME-ResNet and ME-ResNet-KD quit on the first exit, they both require only 4ms to finish the training, which reduces about 10ms compared with the original ResNet. Another interesting phenomenon is that when the model quits on the first exit, the KD technology does not increase the latency compared to the ME-ResNet. When the exit changes from 2 to 7, there is at most  $1$ms added in latency. These results indicate that the multi-exit mechanism can bring flexibility to the computational power-limited devices, and each device can reduce the training time significantly through early exit. Moreover, the latency is marginally increased while applying the KD technology.

Fig. \ref{Fig::model_size} shows the model size in MBytes versus the index of exit of ME-ResNet and the original ResNet, where the number of exits is 7 in ME-ResNet. We can find from this figure that the mode size of the ResNet is 44.6MB, which is similar to 44.7MB of ME-ResNet when the exit is the last one.  However, while the ME-ResNet chooses the earlier exits, its model size  decreases drastically. In particular, when each device chooses the first three exits, the model sizes are all less than 5MB, which is ten times lower than that of the original ResNet. This can significantly reduce the uploading latency and system communication overhead. These results have demonstrated that adding 7 exits for ResNet does not impose a significantly increased burden. More importantly, the ME-ResNet can control its size by exiting from different depths, which enables it to reduce model size as needed. 

Fig. \ref{Fig::acc_vs_rounds} illustrates the test accuracy versus communication rounds, where the dataset is the Fashion-MNIST under the non-IID setting, the total bandwidth $B_A$ is 40MHz, and the latency threshold $\gamma_{th}$ is $15$s. We also compare the convergence performance of the proposed ME-FEEL with different exits to the convergence of FEEL and FEEL-UB in which only one exit is available. This figure shows that the proposed ME-FEEL outperforms the other methods in both the convergence rate and accuracy. In particular, when the $7$-th exit is chosen, the proposed ME-FEEL has a remarkable improvement in the test accuracy compared to FEEL and FEEL-UB. Moreover, the training process of ME-FEEL is more stable than the other methods. In addition, when the $4$-th, $5$-th, and $6$-th exits are chosen, the proposed ME-FEEL still exhibits a clear advantage on the convergence over the other methods. Notably, early exits usually underperform the single exit in FEEL due to the insufficient fitting ability. However, we can find that the early exits in ME-FEEL, such as the first, second, and third one,  still perform better than FEEL and FEEL-UB in the experiments, although with a little degradation compared to its later exits. These results show the effectiveness of the proposed ME-FEEL as well as the exit selection and bandwidth algorithm, which can perform better than the conventional methods by aggregating more updates in each communication round. Besides, it demonstrates that the later exits can transfer knowledge to the earlier ones by the KD technology and help improve the performance.

\begin{table*}[ht]
\begin{threeparttable}
\caption{Accuracy comparison of different methods with $\gamma_{th}=15$s and $B_A=40$MHz.}
\label{tab:acc}
\setlength\tabcolsep{1pt} 

\begin{tabular*}{7.05in}{@{\extracolsep{\fill}} l cccccccc} 
\toprule
     & & & &
     \multicolumn{2}{c}{Accuracy} \\
\cmidrule{2-9}
     Methods & Exit 1 & Exit 2 & Exit 3 & Exit 4 & Exit 5 & Exit 6 & Exit 7 & Maximum\\
\midrule
    FEEL-ideal &  -& -& -&- &- &- &0.87&0.87 \\
    \midrule
     FEEL  &  -& - & -& -  & - & -&0.8079&0.8079 \\
     FEEL-UB & -  &  -& - & -  & - & &0.6542&0.6542 \\
     ME-FEEL (without KD)& 0.7535& 0.7735	& 0.7824& 	0.7975& 	0.8075& 	\textbf{0.8374}	& \textbf{0.8174} &	\textbf{0.8390}\\
     ME-FEEL& \textbf{0.8159} & \textbf{0.8185}&\textbf{0.8178}&\textbf{0.8200}& \textbf{0.8205} & 0.8205&0.8148&0.8286 \\

\bottomrule
\end{tabular*}

\smallskip
\scriptsize
\end{threeparttable}
\end{table*}
In addition, we present a detailed performance comparison on the Fashion-MNIST dataset in Table \ref{tab:acc}, where there are five methods, including FEEL-ideal, FEEL, FEEL-UB, ME-FEEL (without KD) and ME-FEEL. We show the average accuracy  of each available exit and the maximum accuracy of different methods in this table. Except the FEEL-ideal, the communication bandwidth $B_A$ is set to 40MHz and $\gamma_{th}$ is set to 15s for the other four methods. From this table, we can find that the FEEL-ideal can achieve a maximum accuracy of 87\% under the non-IID setting. Notably, there is not any constraint at all here.  However, in the resource-limited environments with aforementioned setting, the performance of the conventional FEEL will be severely degraded. In particular, the FEEL achieves only a maximum accuracy of 80.8\%, which is about 7\% worse than that of the FEEL-ideal. Similarly, the FEEL-UB faces a more serious situation, where the accuracy loss is 22\%. In contrast, the proposed ME-FEEL achieves a maximum accuracy of 84\% without applying the KD technology. These results further verify the effectiveness of the proposed ME-FEEL framework.

Moreover, we can also find from Table. \ref{tab:acc} that there is about $1$\% accuracy loss when we apply the KD technology in ME-FEEL, compared to the ME-FEEL (without KD). However, the early exits of ME-FEEL, including the exits from $1$ to $5$, all perform better than those of ME-FEEL (without KD). 
In particular, the ME-FEEL with the first exit achieves an accuracy gain of 6\%, compared to the method without applying the KD technology. Notably,  ME-FEEL with the first exit also outperforms the conventional FEEL and FEEL-UB with all exits in terms of a lower cost in computation and model size. Similarly, due to the use of the KD technology, the accuracy improvement of the exits from 2 to 5 are 4.5\%, 3.53\%, 2.25\%, and 1.3\%, respectively, compared to the situation of not utilizing the KD technology. This is helpful for the computation-limited devices in the edge-enabled IoT network. The results verify the effectiveness of applying the KD technology into the proposed ME-FEEL.

To further verify the robustness of the proposed ME-FEEL, the performance comparison under different latency thresholds is presented in Fig. \ref{Fig::acc_vs_time}, where the Fashion-MNIST dataset is used, the total communication bandwidth $B_{A}$ is set to $40$MHZ, and latency threshold $\gamma_{th}$ varies from 12s to 15s. We can observe from this figure that the performances of different methods are improved with a larger $\gamma_{th}$. In detail, under the loose latency thresholds such as 21s, the proposed ME-FEEL achieves the accuracy gain of 0.5\% and 5.0\%  compared to that of the FEEL and FEEL-UB, respectively. Moreover, the performance gap enlarges along with the decreased latency threshold. For example, when $\gamma_{th}=15$s and $\gamma_{th}=12$s, compared with the FEEL, the accuracy gain achieved by the proposed ME-FEEL is 1.56\% and 2.68\%, respectively. In addition, the proposed ME-FEEL can obtain the accuracy gain up to 16.6\% and 32.7\% compared to the FEEL-UB. 
These results indicate that the proposed ME-FEEL can adapt well to the practical edge-enabled IoT environments by reasonable exit selection and bandwidth.

Fig. \ref{Fig::acc_vs_bandwidth} shows the accuracy  comparison versus the communication bandwidth $B_A$, where the Fashion-MNIST dataset is used, the latency threshold $\gamma_{th}$ is set to $15$s, and $B_{A}$ varies from 35MHz to 50MHz. 
From this figure, we can find that the accuracy performances of all methods deteriorate along with the decreased $B_A$. However, the performance degradation exhibits different sensitiveness to the decreased bandwidth.  In particular, the accuracy of FEEL-UB deteriorates from 72.7\% to 59.9\%, when the bandwidth varies from 50MHz to 35MHz, indicating that the FEEL-UB is significantly affected by the decrease of total communication bandwidth. The  FEEL performs a little bit better than FEEL-UB, while its accuracy performance is still limited by bandwidth, where the accuracy deteriorates from 80.6\% to 77.9\% when $B_A$ changes from 50Mhz to 35Mhz. Notably, the proposed ME-FEEL outperforms the other methods, and it has a relatively minor performance degradation. In particular, when $B_{A}=50$MHz, the accuracy gain of ME-FEEL is 2.6\% and 10.5\% compared to FEEL and FEEL-UB, respectively. Moreover, the accuracy gain increases when the communication bandwidth decreases, and it is up to 20.75\%. This is because that the proposed ME-FEEL can schedule the system resources efficiently, and thereby it is more robust than the other methods. These results further show that the proposed ME-FEEL can meet the demands of practical edge-enabled IoT networks.

\section{Conclusions}
In this paper, we investigated the problem of facilitating efficient and flexible FEEL in edge-enabled industrial IoT networks.  We proposed a novel FL framework named FEEL to tackle the system heterogeneity problem. The proposed ME-FEEL imposed multiple exits on the local model such that we could avoid training the complete model under the latency constraint. In this case, the devices were able to choose the best exit and train a specific part of the model according to their needs. Since the early exits may cause some performance degradation, we applied the KD technology to solve the problem. Moreover, we proposed a joint exit selection and bandwidth allocation algorithm based on the greedy approach to maximize the expected number of exits in each communication round. Finally,  we conducted simulations to evaluate the performance of the proposed ME-FEEL by employing the Fashion-MNIST dataset with non-IID setting. Simulation results showed that the proposed ME-FEEL could outperform the conventional FEEL in the resource-limited IoT networks.

Nevertheless, there are still some challenges regarding the deployment of computational intelligence and federated learning in edge-enabled industrial IoT networks. For example, how to use the huge number of unlabeled samples in IoT devices and how to train a shared model when the data is presented as a stream still remain challenges. Besides, a more efficient distributed inference protocol needs to be devised to further reduce the latency and energy consumption.


%

\ifCLASSOPTIONcaptionsoff
  \newpage
\fi

\bibliographystyle{IEEEtran}

\bibliography{references}

\end{document}